\documentclass[journal]{IEEEtran}

\usepackage{amsmath}
\usepackage{bm}
\usepackage{algorithm}
\usepackage{color}
\usepackage{cite}
\usepackage{graphicx}
\usepackage{amssymb}
\usepackage{amsmath}
\usepackage{multirow}
\usepackage{subfigure}
\usepackage{epsfig}
\usepackage{framed}
\usepackage{algorithmic}
\usepackage{multirow}
\usepackage{stfloats}
\usepackage{lettrine}
\usepackage{tabularx}
\usepackage{caption}
\usepackage{extarrows}

\begin{document}

\bibliographystyle{ieeetr}

\title{Prognostics Estimations with Dynamic States}
\author{Rong-Jing Bao, Hai-Jun Rong,  ~\IEEEmembership{Member,~IEEE}, Zhi-Xin Yang, ~\IEEEmembership{Member,~IEEE} and Badong Chen, ~\IEEEmembership{Senior Member,~IEEE}
\thanks{Rong-Jing Bao, Hai-Jun~Rong are with the State Key Laboratory for Strength
and Vibration of Mechanical Structures, Shaanxi Key Laboratory of Environment and Control for Flight Vehicle, School of Aerospace, Xi'an Jiaotong University, Xi'an 710049, China (e-mail: brj030@stu.xjtu.edu.cn, hjrong@mail.xjtu.edu.cn).
Zhi-Xin Yang is with the Department of Electromechanical Engineering, Faculty of Science and Technology, University of Macau, Macao SAR 999078, China (e-mail: zxyang@umac.mo).
Badong Chen is with the Institute of Artificial Intelligence and Robotics, School of Electronic and Information Engineering, Xi'an Jiaotong University, Xi'an 710049, China (e-mail: chenbd@mail.xjtu.edu.cn).}
}
\markboth{}%
{}
\maketitle
\begin{abstract}
The health state assessment and remaining useful life (RUL) estimation play very important roles in prognostics and health management (PHM), owing to their abilities to reduce the maintenance and improve the safety of machines or equipment. However, they generally suffer from this problem of lacking prior knowledge to pre-define the exact failure thresholds for a machinery operating in a dynamic environment with a high level of uncertainty. In this case, dynamic thresholds depicted by the discrete states is a very attractive way to estimate the RUL of a dynamic machinery. Currently, there are only very few works considering the dynamic thresholds, and these studies adopted different algorithms to determine the discrete states and predict the continuous states separately, which largely increases the complexity of the learning process. In this paper, we propose a novel prognostics approach for RUL estimation of aero-engines with self-joint prediction of continuous and discrete states, wherein the prediction of continuous and discrete states are conducted simultaneously and dynamically within one learning framework. 
\end{abstract}

\IEEEpeerreviewmaketitle

\section{Introduction}
\lettrine[lines=2]{\textbf{P}}{rognostics} and health management (PHM) has become one of the most important condition-based maintenance (CBM) activities in the aviation industries \cite{MacIsaac2011book, Jardine2006A, Hatem2016Prognostics} with the growing demand for improving the quality and reliability of aircrafts. It mainly focuses on predicting the health state assessment and provides the ability to estimate the remaining life of a particular system or component (aero-engine, bearing, actuator, etc.), which enables intelligent decision making for life-critical and mission-critical applications\cite{Altay2015Prediction, Balaban2015Prognostic, TIE1, TIE2, TIE6, TIE9, TIE14}. Earlier before 2008, the Prognostics Center of Excellence (CoE) at NASA Ames Research Center started the process of extending a test bed that would allow the comparative analysis of different prognostic algorithms. Besides, PHM has already been accepted by the aerospace industry particularly, and the engineering systems community generally, as the direction of today and future in the prognostics research field \cite{AIAA2010Requirements, AIAA2012Requirements, Pecht2009book}.

The common thread among the various avenues of PHM technology development is the estimation of remaining useful life (RUL), which gives operators a potent tool in decision making by quantifying how much time is left until the functionality of a degrading component is lost. Accurate estimation of RUL plays an important role especially in aviation and aerospace systems such that the costs of excessive or insufficient maintenance may be avoided and fatal accidents may be reduced.
According to the researches and literatures, various prognostics approaches including model-based approaches, data-driven approaches and hybrid techniques combining them have been applied to RUL estimation\cite{Schwabacher2005A, Iverson2012General, An2013Options, TIE15, Khan2018a}. The model-based approaches require a priori mathematical and physical knowledge of the process for constructing models. {Nevertheless, it may be very difficult or even impossible to construct accurate physical models for real-world complex systems, which limits the applicability of model-based approaches.}

In recent years, data-driven approaches have been a significant push towards prognostics due to their capability of predicting nonlinear functional and dynamic dependencies. It is feasible for data-driven approaches to automatically characterize and predict the behaviors of a component or system where the monitoring data can be easily observed by sensors to represent the fault propagation trends.
Many data-driven approaches with emphasis on artificial intelligence (neural network, fuzzy system, genetic algorithm, etc.) and statistical learning (hidden Markov model, stochastic process, regression-based model, etc.) have been increasingly applied to the RUL estimation in different areas, such as nuclear system \cite{zio2010a}, machine spindle \cite{TIE4}, tool wear \cite{TII5}, bearing \cite{TIE8}, fatigue-crack-growth \cite{TIE12} and lithium-ion battery \cite{TII6}.
Since the PHM challenge problem with turbofan engine degradation was published by the Prognostics \mbox{CoE} at NASA Ames Research Center in 2008 \cite{C-MAPSS2008data}, more and more researchers around the world have devoted themselves to the development of prognostics, especially based on data-driven approaches \cite{Ramasso2014Performance}.

In this paper, we propose a novel prognostics approach for RUL estimation of aero-engines with self-joint prediction of continuous and discrete states within one learning framework. To achieve this, the quantized kernel recursive least squares algorithm (QKRLS) \cite{Chen2013Quantized} is adopted in our work. The QKRLS belongs to a class of
nonlinear adaptive filtering algorithms derived in reproducing
kernel Hilbert space (RKHS) based on the linear structure of
this space. During the modeling process of the QKRLS, it adopts the self-evolving learning process \cite {rong2006sequential,CEFNS2017} to self-adapt both
the structures and parameters online to capture the dynamical changes of data patterns.
The structure of the QKRLS is related with the kernel centers and identified through an online vector quantization method.
The kernel centers partition the input space into smaller regions. Similar to \cite{Javed2013Novel,Javed2015A}, they can regarded as the fault modes.

\section{Establishment of Predictors}\label{section2}
The main idea of the proposed approach is to build multivariate degradation prognostics models, which are able to achieve the continuous and discrete state prediction of aero-engines simultaneously. The continuous states represent the values of sensor measurements or features while the discrete ones denote the aero-engine health conditions during the degradation. In the following, their self-joint prediction is described in details.

For a training dataset ${\bm T}$ given by \eqref{sample1}, it contains multidimensional signals of $s$ sensors and $t_f$ running cycles in a time series. 
\begin{equation}\label{sample1}
{\bm T}=
\left[
\begin{array}{ccc}
x_1^1 & \cdots &x_{t_f}^1 \\
\vdots & \ddots & \vdots \\
x_1^s & \cdots & x_{t_f}^s \\
\end{array}
\right]_{s \times t_f}
\end{equation}
For establishing a predictor, the training dataset ${\bm T}$ is reorganized in the following form
\begin{equation}\label{datasetT}
\tilde{\bm T}= \{({\mathbf x}_t,{\mathbf d}_t)\}
\end{equation}
where ${\mathbf x}_t = [x_{t-k}^1, \cdots, x_{t-1}^1, \cdots, x_{t-k}^s, \cdots, x_{t-1}^s]^T$ is an $(s \cdot k) \times 1$ dimensional input of the predictor, ${\mathbf d}_t = [x_{t}^1, \cdots, x_{t}^s]^T$ is an $s \times 1$ dimensional output of the predictor and $k$ is the number of regressors.
The modeling of the predictor for continuous states is established as follows:
\begin{equation}\label{model}
\hat{{\mathbf d}}_{t} = P({{\mathbf x}}_{t}, \hat{\bm{\beta}}_{t},{\bm C}_{t})
\end{equation}
where $P$ denotes a nonlinear mapping function, $\hat{\bm{\beta}}_{t}$ and ${\bm C}_{t}$ are respectively the weight parameter and quantization codebook existing in the QKRLS algorithm.

In the QKRLS algorithm, the learning problem of a continuous mapping $P: {\mathbb X} \rightarrow {\mathbb R}$ is regarded as a least squares regression based on a sequence of the observed dataset $\{({\mathbf x}_t, {\mathbf d}_t)|t=1, \cdots, t_f \}$, where ${\mathbb X} \subset {\mathbb R}$ is the input space. According to the Mercer's theorem, any Mercer kernel $\kappa(\cdot, \cdot)$ is able to transform the input space ${\mathbb X}$ to an infinite dimensional  reproducing kernel Hilbert space ${\mathbb F}_\kappa$ (RKHS) by means of a nonlinear mapping $\bm{\phi}$, i.e., $\kappa({{\mathbf x}},{{\mathbf x}}')  = \langle \phi({{\mathbf x}}), \phi({{\mathbf x}}') \rangle_{{\mathbb F}_\kappa}$. In the feature space the inner product $\langle \cdot, \cdot \rangle_{{\mathbb F}_\kappa}$ can be easily computed using the well known kernel trick $\kappa({{\mathbf x}},{{\mathbf x}}') = \phi({{\mathbf x}})^T\phi({{\mathbf x}}')$.
In this paper, the commonly used Gaussian kernel with kernel width $\sigma$ is selected as the Mercer kernel \cite{Chen2013Quantized}.

To achieve the mapping $P$ associated with $\hat{\bm{\beta}}$, one needs to find such a high-dimensional weight $\hat{\bm{B}} \in {\mathbb F}_\kappa$ in the feature space ${\mathbb F}_\kappa$ for the learning problem which minimizes
\begin{align}\label{Q-k-ELM}
\mathop{\min}_{{\bm{B}} \in {\mathbb F}_\kappa}\alpha ||{{\bm{B}}}||_{{\mathbb F}_\kappa}^2 + \sum_{t=1}^{t_f}||{\mathbf d}_t -  {\bm{B}}^T\phi(\text{Q}({{\mathbf x}}_t))||^2
\end{align}
where $\alpha$ is the regularization factor that controls the smoothness of the solution and avoids over-fitting. The feature space ${\mathbb F}_\kappa$ is isometric-isomorphic to the RKHS induced by the kernel. The relationship between $\hat{\bm{B}}$ and $\hat{\bm{\beta}}$ can be easily recognized by the following expression
\begin{equation}\label{Gamma1}
\hat{\bm{B}} = \bm{\Phi}\hat{\bm{\beta}}
\end{equation}
where
$\bm{\Phi} = [\phi({\mathbf x}_{1}), \phi(\bm{\mathbf x}_{2}), \cdots, \phi(\bm{\mathbf x}_{t_f})]$, $\text{Q}(.)$ denotes a vector quantizer (VQ).
As shown in Fig. \ref{figure-model}, the quantization codebook $\bm{C}$ is initially empty and is assumed that it contains $n_L$ code vectors at $t = t_f$, i.e., ${\bm C}_{t_f} = \{\bm{c}_n \in \mathbb{X}\}_{n \in \mathfrak{L}}$,  where $\mathfrak{L} = \{1, 2, \cdots, n_L\}$ is the index set that contains $n_L$ elements. And consequently, the vector quantization operator $\text{Q}({\mathbf{x}}_t)$ maps the input $\{{\mathbf{x}}_t|t = 1, \cdots, t_f\}$ in $\mathbb{X}$ into one of the $n_L$ code vectors in the quantization codebook ${\bm C}_{t_f}$. By partitioning the input space $\mathbb{X}$ into $n_L$ disjoint and exhaustive regions $\Omega_{1}, \Omega_{2}, \cdots, \Omega_{n_L}$, where we define $\Omega_n = \text{Q}^{-1}(\bm{c}_n) \in {\mathbb X}$. The values of the code vectors specify the vector quantization operator $\text{Q}({\mathbf{x}}_t)$, i.e., $\text{Q}({\mathbf{x}}_t) = \bm{c}_n$, if ${\mathbf{x}}_t \in \Omega_n$. Defining $M_n$ is the number of the input data that lies in the $n$th region $\Omega_n$, we have
\begin{equation}
M_n = |\{{\mathbf{x}}_t|{\mathbf{x}}_t \in \Omega_n, 1 \leq t \leq t_f\}|
\end{equation}
$|Z| \ge 0$ denotes the cardinality of a set $Z$ and $\sum_{n=1}^{n_L} M_n = t_f$. Define ${\mathbf d}_{n,i}$ as the desired output that corresponds to the $i$th element from the $n$th region, we have ${\mathbf d}_{n,i} = {\mathbf d}_t$ when ${\mathbf{x}}_t \in \Omega_n$ and $|\{{\mathbf{x}}_{t'}|{\mathbf{x}}_{t'} \in \Omega_n, 1 \leq t' \leq t\}| = i$. Then \eqref{Q-k-ELM} can be rewritten as follows
\begin{align}\label{QKRLS}
\mathop{\min}_{{\bm{B}} \in {\mathbb F}_\kappa}\alpha ||{\bm{B}}||_{{\mathbb F}_\kappa}^2 + \sum_{n\in\mathfrak{L}}\left(\sum_{i=1}^{M_n}||{\mathbf d}_{n,i} -  {\bm{B}}^T\phi(\bm{c}_n)||^2\right)
\end{align}
Consequently, the solution to \eqref{QKRLS} can be derived as follows
\begin{equation}\label{Gamma}
\hat{\bm{B}} = \bm{\Phi} \left( \Lambda \bm{\Phi}^T\bm{\Phi} + \alpha \mathbf{I}\right)^{-1}\bar{\mathbf d}
\end{equation}
where $\bm{\Phi} = [\phi(\bm{c}_{1}), \phi(\bm{c}_{2}), \cdots, \phi(\bm{c}_{n_L})]$, $\Lambda = {\text {diag}}[M_{1}, M_{2},..., M_{n_L}]$
and
\begin{equation}
\bar{\mathbf d} = [\sum_{i=1}^{M_{1}}{\mathbf d}_{1,i}, \sum_{i=1}^{M_{2}}{\mathbf d}_{2,i}, \cdots, \sum_{i=1}^{M_{n_L}}{\mathbf d}_{{n_L},i}]^T
\end{equation}
According to \eqref{Gamma1} and \eqref{Gamma}, we can obtain
\begin{align}\label{beta}
\hat{\bm{\beta}} = \left( \Lambda \bm{\Phi}^T\bm{\Phi} + \alpha \mathbf{I}\right)^{-1}\bar{\mathbf d} = \left(\Lambda \Psi + \alpha \mathbf{I}\right)^{-1}\bar{\mathbf d}
\end{align}
where $\Psi = \bm{\Phi}^T \bm{\Phi}$ is the Gram matrix and its elements are $\Psi_{ij} = \kappa(\bm{c}_i, \bm{c}_j)$.

When a predictor $P(\hat{\bm{\beta}}, {\bm C})$ of \eqref{model} is established, the codebook $\bm{C} =\{{\bm c}_n|n =1, 2, \cdots, n_L\}$ is obtained accordingly. The multivariate degrading signals of the training dataset $\{{\mathbf x}_t|{\mathbf x}_t \in \mathbb{R}^{s \cdot k}, t = 1, \cdots, t_f\}$ can be partitioned into $n_L$ regions, where ${\mathbf x}_t$ belongs to the $n$th region $\Omega_n$ by calculating
\begin{equation}\label{state_omega}
n = \mathop{\text{arg min}}_{1 \leq n \leq n_L}||{\mathbf{x}}_{t} - {\bm c}_n||
\end{equation}
Each region $\Omega_n$ corresponds to a code vector $\bm{c}_n$ as its center and is considered as a discrete state of an engine. 
The first region $\Omega_1$ indicates the initial state where the engine starts. Relatively, the final region $\Omega_{n_L}$ represents the failure state where the engine comes to end-of-life (EoL). It should be noted that all the regions can be viewed as the transition from normal states to degrading states until the final fault state. Thus, the discrete states are equivalent to the codebook $\bm C$. When the codebook $\bm C$ is obtained, the discrete states representing the failure thresholds are determined.

\section{RUL Estimation}\label{section3}
For a testing dataset ${\bm S}$ given by
\begin{equation}\label{sample2}
{\bm S}=
\left[
\begin{array}{ccc}
{\tilde x}_1^1 & \cdots &{\tilde x}_{t_c}^1 \\
\vdots & \ddots & \vdots \\
{\tilde x}_1^s & \cdots & {\tilde x}_{t_c}^s \\
\end{array}
\right]_{s \times t_c}
\end{equation}
the multidimensional signals are observed by $s$ sensors when an engine operates normally up to $t_c$ running cycles before system failure. By reorganizing the testing dataset ${\bm S}$ as well as \eqref{datasetT} in the following form
\begin{equation}
\tilde{\bm S}= \{({\tilde{\mathbf x}}_t,{\mathbf d}_t)\}
\end{equation}
where ${\tilde{\mathbf x}}_t = [{\tilde x}_{t-k}^1, \cdots, {\tilde x}_{t-1}^1, \cdots, {\tilde x}_{t-k}^s, \cdots, {\tilde x}_{t-1}^s]^T$ and
${\mathbf d}_t = [{\tilde x}_{t}^1, \cdots, {\tilde x}_{t}^s]^T$.

Prior to the RUL estimation of a testing engine, its future degrading signals $\{\hat{{\mathbf d}}_{t}|t > t_c\}$ can be estimated by the predictor
\begin{equation}\label{prediction}
\hat{{\mathbf d}}_{t} = P(\hat{\bm{\beta}},{\bm C}, {\tilde{{\mathbf x}}}_{t}) = [\hat{\tilde x}_{t}^1, \cdots, \hat{\tilde x}_{t}^s]^T
\end{equation}
where $ t = t_c+1, t_c+2, \cdots$. It can be seen that the prediction starts from $t = t_c+1$ to predict the future degrading signals $[\hat{\tilde x}_{t_c+1}^1, \cdots, \hat{\tilde x}_{t_c+1}^s]^T$ in time sequence. Particularly, when the time sequence $t \in [t_c+2, t_c+k]$, the inputs contain the observed data ${\tilde x}^i_{t}$ and the estimated data $\hat{\tilde x}^i_{t}$. And when the time sequence $ t > t_c+k+1$, the future degrading signals are all predicted by the estimated ones $\hat{\tilde x}^i_{t}$. 

For testing samples comprised of $\mathfrak N$ time series, each time series is also from a different testing engine of the same fleet which has a diverse initial condition and operational environment. we need to determinate the optimum predictor for each time series such that an accurate RUL estimation can be obtained.
For this purpose, two criteria are adopted here. The first one is named as the RMSE (Root Mean Square Error) criterion which is utilized to determinate the predictor for each specific testing dataset. It guarantees that the future degrading signals can be predicted precisely.
Based on the RMSE criterion, appropriate predictors are determinated through calculating the testing RMSE which is given as follows
\begin{equation}\label{RMSE1}
{\rm RMSE}_i = \sqrt{\sum_{t=1}^{t_c}||{{\mathbf d}}_t - \hat{{{\mathbf d}}}_{t}^i||^2}, i = 1,..., \mathfrak M
\end{equation}
where $t \leq t_c$, ${\mathbf d}_{t}$ is the desired output, $\hat{{{\mathbf d}}}_{t}^i = P^i(\hat{\bm{\beta}}^i,{\bm C}^i, {\tilde{\mathbf x}}_{t})$ is the output of the predictor, $\hat{\bm{\beta}}^i$ and $C^i$ are the weight parameter and codebook of the predictor $P^i$ respectively.
According to the testing errors, the first $J$ predictors with smaller errors $\{P^j|j = 1,...,J < \mathfrak M\}$ are selected under the RMSE criterion.

Generally, when an engine fails to run or arrives at the fault mode at the failure time $t_f$, the degrading signals grow to the damage level.
For a testing engine running up to the current time $t_c$ before system failure, its health states of future degradation are unknown to the users.
If the testing engine continues to operate and is degrading increasingly due to some wear and tear, the states of degradation will terminate at the fault state $\Omega_{n_L}$.
The discrete states are considered as different degrading levels of the degradation process.

In this study, the discrete states are determinated during the predicting process simultaneously and directly rather than by using the classifiers or clusters. The codebook of a predictor represents the states of an engine from healthy condition to fault mode, which is decided and computed through a vector quantizer. While the future degrading signals of a testing engine are predicted with its optimum predictor $P^{j^*}$, the discrete states can be determinated accordingly by the codebook ${\bm C}^{j^*}$ of $P^{j^*}$. If the predicted multidimensional signals ${\tilde{\mathbf x}}_t$ belong to the final region $\Omega_{n_L}$ of $P^{j^*}$, the prediction of the testing engine stops. Once the predicted multidimensional signals ${\tilde{\mathbf x}}_t$ come into the final region $\Omega_{n_L}$, the failure time $t_f$ of this testing engine can be calculated by using the following distance metric
\begin{equation}\label{tf}
t_f = {\text {min}} \left(\mathop{\arg\min}_{{\text {s.t.}}\  t > t_c}||{\tilde{\mathbf x}}_{t} - {\mathbf c}_{n_L}||\right)
\end{equation}
And the RUL of this testing engine is estimated conveniently between the current time $t_c$ and the failure time $t_f$
\begin{equation}\label{estimated RUL}
\hat{t}_{RUL} = t_f - t_c
\end{equation}

\section{Application and Results}\label{section4}
\subsection{Datasets and Evaluation Metrics}
In order to demonstrate the performance of the proposed prognostics approach, three text files ``train-FD001.txt'', ``test-FD001.txt'' and ``RUL$\_$FD001.txt'' of the PHM 2008 challenge datasets \cite{C-MAPSS2008data} are employed here, wherein one condition and one fault mode are considered. 
In this paper, five predictable sensors $[2, 8, 11, 13, 15]$ (7th, 13th, 16th, 18th and 20th column of the dataset) are employed for RUL estimation by a data-mining technique\cite{Javed2015A}.

As in \cite{Ramasso2013Joint, Khelif2014RUL, Ramasso2014Investigating-b, Javed2015A, Malhotra2016Multi, TIE3, Gugulothu2018Predicting}, the following evaluation metrics are used for PHM performance evaluation such as coefficient of determination (R2), RUL Error PDF, mean square error (MSE), mean absolute error (MAE), mean absolute percentage error (MAPE) Score, Accuracy Rate. Their details can be referred to the above references.

\subsection{Simuation Results and Comparison}
\begin{table*}[!htbp]
\scriptsize
\renewcommand\arraystretch{1.05}
\setlength{\abovecaptionskip}{0pt}
\setlength{\belowcaptionskip}{-2pt} \centering \caption{COMPARISON RESULTS BETWEEN DIFFERENT APPROACHES FOR 100 TESTING ENGINES}
\begin{tabularx}{1\textwidth}{@{\hspace{0.1cm}}c@{\hspace{0.35cm}}c@{\hspace{0.3cm}}c@{\hspace{0.3cm}}c@{\hspace{0.3cm}}c@{\hspace{0.3cm}}c@{\hspace{0.3cm}}c@{\hspace{0.3cm}}c@{\hspace{0.3cm}}c@{\hspace{0.1cm}}}
\hline\hline
Results$\setminus$Approach          &EVIPO-KNN \cite{Ramasso2013Joint}&IBL  \cite{Khelif2014RUL}&	RULCLIPPER \cite{Ramasso2014Investigating-b}&SWELM-SMEFC \cite{Javed2015A}   &LR-ED$_2$ \cite{Malhotra2016Multi}&  SVR \cite{TIE3}&Embed-LR$_1$ \cite{Gugulothu2018Predicting}&	Self-Predciton (Proposed) \\ \hline
R2	                 &/	               &/         &	/	               &0.614	   &/             &/	       &/	                &\textbf{0.911}\\
RUL Error PDF &[-85,120]       &/	   &/	                &[-39,60]	    &/	       &/         &/	                &[-53,43]\\
In time 	            &53                 &54     &67                   &48             &67          &70       &59               &\textbf{78}\\
Early (FN)	     &36	                &18     &44                    &40             &13          &/         &14                &16 \\
Late (FP)	     &11	                &28	  &56                    &12	          &20	       &/	       &27	          &6\\
MSE	           &/	                &/	  &176                  &/	          &164        &/	       &155	          &\textbf{153.7}\\
MAE	           &/                    &	/	  &10	                &/	          &9.9         &/	       &9.8	          &\textbf{7.34}\\
MAPE             &/	               &/	         &20\%	         &/	          &18\%      &/        &19\%	          &\textbf{9.95\%}\\
Score              &/	               &/	         &\textbf{216}   &1046	         &256         &448.7	&219              &351.6\\
Accuracy Rate	    &53\%             &	54\% &67\%	         &48\%	   &67\%      &70\%	&59\%            &\textbf{78\%}\\
\hline\hline\\
\end{tabularx}
\label{results}
\end{table*}
We give the results of the RUL estimation for total 100 testing engines, which are summarized in Table \ref{results}. The results are also compared to other popular prognostics approaches, such as EVIPO-KNN \cite{Ramasso2013Joint}, IBL \cite{Khelif2014RUL}, RULCLIPPER \cite{Ramasso2014Investigating-b}, SWELM-SMEFC \cite{Javed2015A}, LR-ED$_2$ \cite{Sateesh2016Deep}, SVR \cite{TIE3} and Embed-LR$_1$ \cite{Gugulothu2018Predicting}. The results of these references are arranged in ascending order of the published year and are selected according to their best performances.
Compared to SWELM-SMEFC, the coefficient of determination $R2$ we obtained is larger. Although the larger value of $R2$ stands for better prediction, it is primarily designed for model selection using the training dataset. Thus, we suggest avoiding the use of this metric as a main tool for performance evaluations.
The RUL error distribution of the proposed approach has the lower span $I = [-53, 43]$ than EVIPO-KNN and is similar with SWELM-SMEFC $I = [-39, 60]$. This demonstrates that the dynamic discrete states used in the proposed approach and SWELM-SMEFC are more reasonable and practical than fixed ones in EVIPO-KNN to represent the health of different degrading engine units. In IBL, the number of neighbors used in the k-nearest train instances for RUL estimation is also pre-defined according the best results.

In terms of in time and late prediction, the proposed approach can achieve the best accuracy with 78 in time prediction and the least late prediction that is only 6 among these approaches. From the aspect of score, the proposed approach is comparable with the LR-ED$_2$, the Embed-LR$_1$ and the RULCLIPPER. It should be noted that
 the best performance with respect to score in RULCLIPPER is obtained by using various combinations of features to choose the best set of features. In terms of other evaluation metrics, the proposed approach achieves the best performance with MSE = 153.7, mean absolute error MAE = 7.34 and MAPE = 9.95\% compared to other prognostics approaches. From the above discussions, it is evident that the results obtained in this study is very competitive compared to other prognostics approaches.
Generally, it is quite challenging for prognostics to achieve the continuous states of degrading signals when prior experience or information about degradation process is not available. As a matter of fact, in this proposed approach, the degradation of aero-engines can be self-predicted by combining the prediction of continuous states and discrete states simultaneously. Besides, the RUL estimation is achieved with high accuracy and low score.

\section{Conclusions}\label{section5}
In this paper, we propose a novel prognostics approach with self-joint prediction of continuous and discrete states and apply it in the RUL estimation of aero-engines.
In the proposed approach, the degradation process is evaluated based on the continuous and discrete states. The continuous states depict the values of degrading signals and the discrete ones represent the fault modes. The discrete states indicated by the kernel centers are decided dynamically according to the self-evolving learning process and can be automatically achieved for different engines which degrade diversely according to their operational conditions and environments. This avoids the pre-assumed fault modes before hand and improves the RUL estimation accuracy. Within the QKRLS learning framework, the predictors used to predict the future degrading signals are constructed jointly based on the self-evolving learning. This is different from previous approaches in which the predictors and fault modes are decided separately and this further reduces the complexity of the RUL estimation process.


\end{document}